\documentclass[3p]{elsarticle}

\usepackage{lineno,hyperref}
\modulolinenumbers[5]

\usepackage{graphicx}
\usepackage{amsmath}            
\usepackage{amssymb}            
\usepackage{booktabs}
\usepackage[table]{xcolor}      

\journal{Neural Networks}

\biboptions{authoryear}

\begin{document}

\begin{frontmatter}

\title{Surface Similarity Parameter: A New Machine Learning Loss Metric for Oscillatory Spatio-Temporal Data}

\author[DYN]{Mathies Wedler\corref{mycorrespondingauthor}}
\ead{mathies.wedler@tuhh.de}
\ead[url]{www.tuhh.de/dyn}

\author[DYN]{Merten Stender\corref{cor2}}
\author[DYN]{Marco Klein\corref{cor3}}
\author[DYN]{Svenja Ehlers\corref{cor4}}
\author[DYN,IC]{Norbert Hoffmann\corref{cor5}}

\cortext[mycorrespondingauthor]{Corresponding author}

\address[DYN]{Hamburg University of Technology, Dynamics Group, Schlossm{\"u}hlendamm 30, 21073 Hamburg, Germany}
\address[IC]{Imperial College London, Department of Mechanical Engineering, 58 Prince's Gate, South Kensington, London SW7 1AY, United Kingdom}

\begin{abstract}
Supervised machine learning approaches require the formulation of a loss functional to be minimized in the training phase. Sequential data are ubiquitous across many fields of research, and are often treated with Euclidean distance-based loss functions that were designed for tabular data. For smooth oscillatory data, those conventional approaches lack the ability to penalize amplitude, frequency and phase prediction errors at the same time, and tend to be biased towards amplitude errors.
We introduce the surface similarity parameter (SSP) as a novel loss function that is especially useful for training machine learning models on smooth oscillatory sequences. Our extensive experiments on chaotic spatio-temporal dynamical systems indicate that the SSP is beneficial for shaping gradients, thereby accelerating the training process, reducing the final prediction error, increasing weight initialization robustness, and implementing a stronger regularization effect compared to using classical loss functions. The results indicate the potential of the novel loss metric particularly for highly complex and chaotic data, such as data stemming from the nonlinear two-dimensional Kuramoto–Sivashinsky equation and the linear propagation of dispersive surface gravity waves in fluids. 
\end{abstract}


\end{frontmatter}


\section{Introduction}
\label{sec:introduction}
Given the fact that almost any natural process is dynamic and non-stationary, sequential data is ubiquitous across many scientific disciplines, such as weather forecasting, finance, natural language processing, acoustics and many more. Besides classical analytical and numerical techniques \citep{Anderson.1995, Wriggers.2008}, machine learning approaches have been adopted lately for various regression or classification tasks on sequential data given a continuous quantity $y\left(t\right)$ sampled at discrete instants $\mathbf{y}\left(t\right)= \left[y(t_1), \dots y(t_N) \right]$ \citep{Bagnall_2016, zheng2016, shi2018, Herzog2018-du, vlachas2018, Fawaz2019, leegard2021, Stender_2021}. Most data-driven approaches are rooted in a supervised optimization framework for a model's $f$ parameters $\mathbf{\theta}$, essentially targeting at minimizing an error functional $J\left(\mathbf{y}, \hat{\mathbf{y}}(\theta) \right)$ that describes the discrepancy between the model prediction $\hat{\mathbf{y}}$ and the ground truth $\mathbf{y}$ \citep{LeCun_2015, singh2016}. Some techniques utilize sequence featurization \citep{Fulcher.2014}, i.e. turning sequential data into vectors of descriptive scalar values, while direct approaches perform the error computation on the raw sequences themselves \citep{Hochreiter_1997, Fawaz2019}. There is a plethora of metrics for measuring the similarity of two (multivariate) sequences, such as generic time domain distance metrics or custom metrics tailored towards specific aspects of the given data \citep{Mueller2007, Fulcher_2013, Jiang_2019}. The choice of a specific error metric as loss function for the training of machine learning methods is of crucial importance: the loss function shapes the gradient \citep{LeCun_2015}, thus the convergence behavior of the optimization, and potentially biased predictions. 

Conventional Euclidean distance metrics such as the mean squared error (MSE) and mean absolute error (MAE) do not take the sequential data characteristic into account, nor smoothness or continuity of the predicted sequences. Instead, considering sequences as a set of values sampled from some distribution, the MSE and MAE aim at fitting the mean and median, respectively, of that distribution. Often times, a spectral decomposition of complex (spatio-) temporal dynamics is utilized to describe the data in a finite series of harmonics, each described by amplitude, frequency and phase \citep{Giannakis_2019}. From this viewpoint, Euclidean distance metrics are biased towards amplitude errors, rendering them as sub-optimal loss functions for some machine learning tasks as our work will illustrate. We introduce the surface similarity parameter SSP \citep{Perlin2014ARQ} as a novel loss metric, that is capable of combining amplitude, frequency and phase errors into a single scalar error metric. Thereby, the SSP can provide several benefits for machine learning applications on complex oscillatory sequential data.

Studying the highly chaotic dynamics of the a) one- and b) two-dimensional Kuramoto–Sivashinsky equation (KS-eq.), and the c) propagation of dispersive surface waves of a very broad spectrum, our work indicates that the SSP is a particularly beneficial loss function the more complex the sequences are, that is, the wider the underlying wave spectrum becomes (i.e. the larger the number of relevant harmonics in the spectral decomposition). Our numerical experiments indicate that the SSP enhances optimizer convergence through effective gradient shaping, increases robustness against broad-banded spectra and weight initialization, and introduces strong regularization effects. Further, the benefits of the SSP loss function are shown to be rooted in an error saturation effect, that is effectively bounding the degree to which two sequences can be dissimilar, thereby forming sharper global minima in a flattened error functional in the parameter optimization space.

This work is structured as follows: Section~\ref{sec:method} introduces the surface similarity parameter, re-visits major considerations on machine learning loss functions and briefly introduces the experimental settings. Section~\ref{sec:results} illustrates several in-depth studies comparing different loss metrics for learning a convolutional encoding-decoding flow map for the KS-eq. and a broad-banded dispersive wave spectrum.

\section{Methods}
\label{sec:method}
We will introduce the SSP first, then compare against conventional Euclidean distance metrics, and finally highlight the property of an upward bounded loss surface of the proposed loss function. Subsequently, we briefly present the experimental settings for evaluation of the SSP as a machine learning loss function.

\subsection{Surface similarity parameter}
\cite{Perlin2014ARQ} introduce the surface similarity parameter (SSP) as a quantitative comparison metric for two surfaces $\mathbf{y}_1, \mathbf{y}_2$ with spatial and/or temporal extent in one or more dimensions. In this work, we only consider surfaces with spatial extent, such that the SSP formulation can be written as
\begin{linenomath*}
\begin{equation}\label{eq:ssp}
    J_{\mathrm{SSP}}(\mathbf{y}_1,\mathbf{y}_2)=\frac{\sqrt{\int|F_{\mathbf{y}_1}(\mathbf{k}) - F_{\mathbf{y}_2}(\mathbf{k})|^2d\mathbf{k}}}{\sqrt{\int|F_{\mathbf{y}_1}(\mathbf{k})|^2d\mathbf{k}} + \sqrt{\int|F_{\mathbf{y}_2}(\mathbf{k})|^2d\mathbf{k}}}\in[0,1]
\end{equation}
\end{linenomath*}
where $F_{\mathbf{y}}$ denotes the Fourier transform of the signal $\mathbf{y}(\mathbf{x}, t)$. $\mathbf{k}$ denotes the wave number vector $\mathbf{k}=\begin{bmatrix}k_x\end{bmatrix}$ for one-dimensional or $\mathbf{k}=\begin{bmatrix}k_x, k_y\end{bmatrix}^\intercal$ for two-dimensional spatial inputs, respectively. The SSP is based on the $L^2$ norm, originally denoted as a special case of the Sobolev norm by \cite{Perlin2014ARQ}, of the individual signals $||\mathbf{y}_i||^2=\int|F_{\mathbf{y}_i}(\mathbf{k})|^2d\mathbf{k}$ and their difference in space domain $||\mathbf{y}_1 - \mathbf{y}_2||^2=\int|F_{\mathbf{y}_1}(\mathbf{k}) - F_{\mathbf{y}_2}(\mathbf{k})|^2d\mathbf{k}$. By effectively measuring the similarity of two signals in Fourier space, the SSP inherently penalizes amplitude-, frequency- and phase errors uniformly in a single quantity\footnote{Similar to the A-weighting of sound pressure levels, the SSP may be tuned towards a non-balanced weighting of amplitude-, frequency- and phase errors by imposing a weighting function on the Fourier spectra $F_{\mathbf{y}_i}$}. If both signals are identical, the $L^2$ norm of their difference is 0, such that $\mathrm{SSP}=0$ denotes perfect agreement among the signals. With increasing difference between $\mathbf{y}_1$ and $\mathbf{y}_2$, the SSP approaches 1. In other words, the SSP is a normalized measure for the relative error between two signals, which greatly promotes interpretability. It is noteworthy that $\mathrm{SSP}=1$ (\cite{Perlin2014ARQ} refer to $\mathrm{SSP}=1$ as \emph{perfect disagreement}) is only achieved for two phase-inverted signals disregarding their respective amplitudes. While the SSP definition by \cite{Perlin2014ARQ} utilizes the continuous Fourier transform, we compare discrete signals sampled at the same rate and thus use the discrete Fourier transform (DFT) computed by the fast Fourier transform algorithm (FFT) \citep{CooleyTukey}. By relying on the FFT algorithm, we impose two requirements on signals compared with SSP: While it is mandatory that the signals have enough grid points to include enough harmonic components in Fourier space to faithfully represent the original signals, the performance of the SSP calculation improves drastically when signals sampled at $2^n,\;n\in\mathbb{N}$ grid points are compared.

The SSP is applied in the field of ocean engineering to quantify numeric ocean wave prediction \citep{Klein2019} and -reconstruction methods \citep{desmars_reconstruction_2021} as well as for ensuring consistent conditions in ship movement experiments \citep{van_essen_influence_2021}. Other applications include the quantification of denoised GPS signals \citep{sharie_determination_2020} and demodulated fringe pattern \citep{hernandez-lopez_parallel_2021}. Despite occasional usage in a variety of fields, no publication reports an in-depth analysis on why the SSP is preferred over established Euclidean distance metrics in those fields. Furthermore, and to the knowledge of the authors, the SSP has not been used as a loss metric for machine learning-related research yet. The following paragraphs will highlight conceptual differences of the SSP and Euclidean error metrics, and point out aspects that render the SSP a preferable loss function for machine learning purposes.

\subsection{Is there a limit to dissimilarity?}
Focusing at the predicted model output and the ground truth for long sequential data, a loss function is all about a condensation of many pieces of information into a scalar and abstract deviation quantity. In simple words, the loss function tells \emph{how similar and how dissimilar can two sequences be?} The former case is trivial to answer by almost any Euclidean distance metric. The latter, however, is nontrivial: what is the most \emph{dissimilar} signal $y(t)$ to a given non-zero signal $x(t)$? Euclidean metrics would answer that question by increasing the amplitudes of $y(t)$ towards infinity. Alternatively, one could state that $y(t) = -x(t)$ is the most dissimilar signal to $x(t)$, and what about $y(t)=0$? Looking at Euclidean distance metrics, we can observe that they are un-bounded, and hence over-sensitive to amplitude errors and under-sensitive to phase or frequency errors if we consider signals composed of harmonics. On the contrary, the SSP is a bounded quantity, which is most conveniently shown by calculating the SSP for amplitude scaling signals, i.e. $\mathbf{y}_1$ and $\mathbf{y}_2 = \kappa\cdot\mathbf{y}_1,\;\kappa\in\mathbb{R}$. For this case, Eq.~\eqref{eq:ssp} can be simplified to the $C_0$ continuous function
\begin{linenomath*}
\begin{equation}
    J_{\mathrm{SSP}}(\mathbf{y}_1, \kappa\cdot\mathbf{y}_1) = \frac{|1 - \kappa|}{1 + |\kappa|}\begin{cases}\in[0, 1)&\forall\kappa\in\mathbb{R}^+\\=1&\forall\kappa\in\mathbb{R}_0^{-}\end{cases}, \quad \frac{\mathrm{d}^2J_{\mathrm{SSP}}}{\mathrm{d}\kappa^2} = \begin{cases}>0\;(\mathrm{convex})&\forall\kappa\in [0,1)\\<0\;(\mathrm{concave})&\forall\kappa>1\end{cases}
    \label{eq:bound_ssp}
\end{equation}
\end{linenomath*}
i.e. we observe the SSP running into an upper limit of $\mathrm{SSP}=1$ for $\kappa\in\mathbb{R}_0^-$, meaning that there is a bound to the dissimilarity of two signals when measured in SSP. When used as a ML loss function, this property will flatten the loss surface in remote regions of the parameter space. The loss function shapes the gradients seen by the ML optimizer and thus directly affects the training behavior such as convergence speed and final model accuracy. In general, gradient-based optimization methods like SGD \citep{Robbins1951, Kiefer1952} benefit from large gradients as the step size in parameter space is directly linked to the gradient of the loss surface. More advanced optimization schemes like AdaGrad \citep{duchi2011adaptive}, AdaDelta \citep{Zeiler2012}, RMSprop \citep{hinton2018} or Adam \citep{Kingma.2014} introduce momentum, i.e. information about past gradients. This allows the optimizer to \emph{accelerate} under certain circumstances. \citet{Kingma.2014} state that Adam's step size in parameter space reaches an upper bound ``when a gradient has been zero at all timesteps except at the current timestep", i.e. for plateaus in the loss surface. Purely gradient-based optimizers would be trapped for $\nabla_{\boldsymbol{\theta}}J(\boldsymbol{\theta})=\boldsymbol{0}$, the Adam optimizer on the other hand even benefits from small gradients as shown in \ref{sec:toy_example}. When training a ML model, the initial weights $\boldsymbol{\theta}_0$ are sampled from a certain distribution (we refer to \citet{Narkhede_2021} for a comprehensive overview on weight initialization methods), i.e. the optimization is started at a random location in parameter space. When working with signals composed of harmonics, the prediction of the randomly initialized model will almost certainly include amplitude-, phase- and frequency errors at the same time. In consequence, the optimizer will never encounter a perfectly flat plateau in the loss surface when using the SSP but rather very small gradients $\nabla_{\boldsymbol{\theta}}J(\boldsymbol{\theta})\approx\boldsymbol{0}$.

\subsection{Comparison against Euclidean distance metrics}\label{sec:ssp_vs_euclidean}
\begin{figure}[!h]
    \centering
    \includegraphics[scale=1]{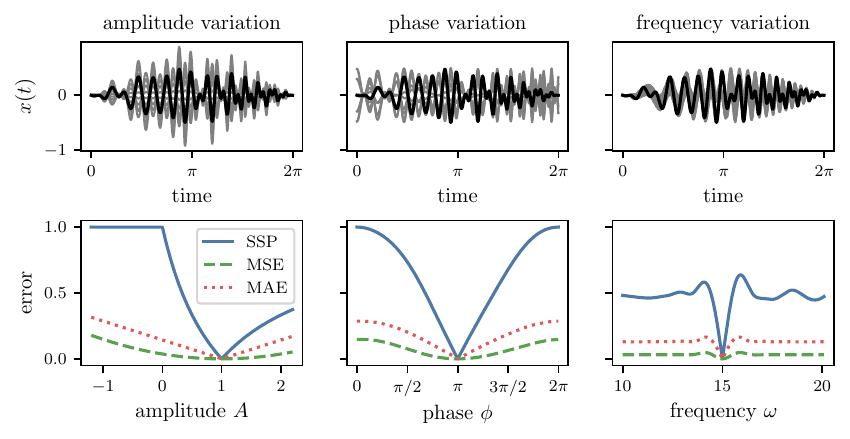}
    \caption{Evaluation of three different distance metrics under amplitude, phase and frequency variation of the sample signal $x(t) = A \cdot (0.5 \sin(0.5 t + \phi) \cdot (0.5 \cos((2 + 3t) t) + 0.5 \cdot \cos(\omega t))$ (gray lines) compared against the baseline signal (solid black line in top panel) defined by $A=1$, $\phi=\pi$, $\omega = 15$}
    \label{fig:bounded_loss}
\end{figure}
Figure~\ref{fig:bounded_loss} displays the MSE and MAE metrics in comparison to the SSP for a synthetic test signal composed of three arbitrary harmonic components including a variable phase term $\phi$, a variable frequency term $\omega$ as well as an amplitude scaling envelope term $A$. Looking at the frequency variation about the reference configuration, there is no qualitative difference among the metrics, particularly not in the gradients of the metrics in vicinity of the optimum. In contrast, this observation does not hold for the amplitude variation. Per definition, the MAE exhibits a constant gradient and the MSE shows a linearly decreasing gradient, i.e. becoming very flat, towards the optimum. The SSP is convex to the left, and concave to the right of the optimum according to Eq.~\eqref{eq:bound_ssp}. While the Euclidean metrics increase monotonously for positive and negative deviations from the reference amplitude, the SSP exhibits the previously discussed upper limit of $\mathrm{SSP}=1$ for zero and negative amplitudes, as well as for larger positives amplitudes. There exists a large and perfectly flat error functional for negative amplitudes $A$, hence forcing the optimizer to take large steps and escape that desert of zero gradients. Once the optimizer approaches the optimum neighborhood, suddenly gradients become large towards the minimum. Similarly, if the true solution is approached from the far right, i.e. positive $A$ values, the error becomes steeper as the minimum is approached. This is not the case for the convex MSE, which becomes flat towards the minimum.

When training a ML model on oscillatory data, the prediction most certainly includes amplitude-, phase- and frequency errors at the same time. Starting from some remote region in the model's parameter space, the loss function is minimized by iterative weight updates. Since Euclidean distance metrics are over-sensitive to amplitude errors, MSE (and MAE) will first fit the mean (and median) amplitudes of $\boldsymbol{y}$ and $\hat{\boldsymbol{y}}$, while phase- and frequency errors play a minor role compared to amplitude. The SSP on the other hand is stricter as it only decreases once a weight-update tackles amplitude-, phase- and frequency errors at the same time. This property will most likely translate into less local minima in the SSP loss surface compared to MSE and MAE. In other words, the SSP is more robust against weight initialization effects as it squashes the error functional and thus local minima in remote regions. One may even conclude that the strictness of the SSP leads to less non-convex regions in the loss landscape.

\subsection{Experimental settings}
\begin{figure}[!h]
    \centering
    \includegraphics[scale=1]{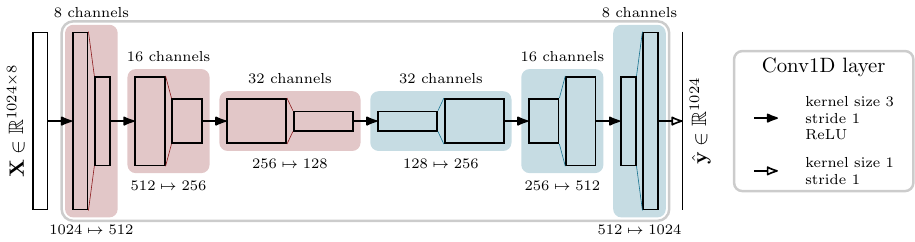}
    \caption{Fully convolutional network with encoder-decoder structure for multivariate sequential data. The model is build from three encoder- (red) and decoder blocks (blue), each of which applies a convolution and a pooling operation. The number of channels is stated above, the pooling operation is stated below each block. By alternating convolution and up (or down-) sampling, the model is able to extract features on different scales. Overall, the model maps an input $\mathbf{X}\in\mathbb{R}^{1024\times 8}$ to an output $\hat{\mathbf{y}}\in\mathbb{R}^{1024}$. For the two-dimensional evaluation cases, the respective two-dimensional instances of the layers depicted here are used, e.g. Conv1D is replaced by Conv2D}
    \label{fig:network}
\end{figure}

For the numerical experiments, we evaluate the performance of a fully convolutional encoder-decoder model learning a flow map, i.e. taking a number of $n_{\tau}=8$ states at preceding time steps and making a prediction about the system state at the next time step. As illustrated in Figure~\ref{fig:network}, the model consists of an encoding path with three stages and a symmetric decoding path. Each convolutional layer in the encoding path is followed by downsampling along the spatial dimension via Max Pooling (upsampling via transposed convolutions in the decoder path), such that we extract features on different scales. While the spatial dimension of the internal input representation is bisected, the amount of convolutional filters is doubled per stage (divided in halve in the decoding path). We apply a final convolutional layer with a filter size of 1 to output $\hat{\mathbf{y}}\in\mathbb{R}^{1024}$. The Adam optimizer \citep{Kingma.2014} is employed with learning rate $\alpha=1\mathrm{e}{-3}$, $1^{\text{st}}$ order exponential decay $\beta_1=0.9$, $2^{\text{nd}}$ order exponential decay $\beta_2=0.999$ and $\varepsilon=1\mathrm{e}{-7}$. For the SGD optimizer \citep{Robbins1951, Kiefer1952}, a learning rate $\alpha=1\mathrm{e}{-2}$ is set with no momentum. 

\begin{figure}[h!]
\centering
    \includegraphics[scale=1]{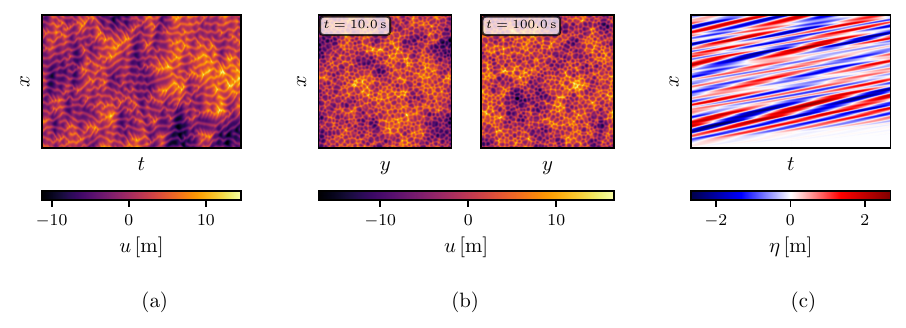}
    \caption{Qualitative samples for each numerical test case: (a) one-dimensional KS-eq., (b) two snapshots ($t=\{10, 100\}\,$s) of the two-dimensional KS-eq. and (c) one-dimensional broad-banded surface waves}
    \label{fig:cases}
\end{figure}

As typical test cases, we study the chaotic spatio-temporal dynamics of the (a) one- and (b) two-dimensional Kuramoto–Sivashinsky equation (KS-eq.) \eqref{eq:KS_equation} and the (c) time evolution of dispersive surface waves. Figure~\ref{fig:cases} shows one qualitative sample of each test case. Cases (a) and (b) have been studied as benchmark cases in many recent works on data-driven forecasting methods \citep{Pathak2017, Pathak.2018, Raissi.2018, vlachas2020}. Importantly, we note that our aim is \emph{not} to propose an optimal neural architecture for future state forecasting, but to study the training behavior of a given model architecture under different loss functions. The training behavior of the given neural architecture is studied under different choices of the loss function $J\in\mathbb{M}=\{\mathrm{SSP},\mathrm{MSE},\mathrm{MAE},\mathrm{RMSE},\mathrm{Huber}\}$, RMSE: Root Mean Squared Error, Huber: Huber loss \citep{Huber1964} with $\delta=1$. All metrics in the set $\mathbb{M}\setminus\{J\}$ are used as error metrics to observe the training behavior in their respective quantity, allowing us to evaluate a model trained on loss function $J$ in terms of all metrics in $\mathbb{M}$. To cross-validate our results and analyze their variability, each loss function $J\in\mathbb{M}$ is used to train the model on 50 pseudo random data splits: The set of all preprocessed samples $(\mathbf{X}_i\in\mathbb{R}^{1024\times8}, \mathbf{y}_i\in\mathbb{R}^{1024})$ is deterministically shuffled using the training count $j\in\{1, 2, ..., 50\}$ as random seed and then split with ratio 70/30 into training- and validation set. The preprocessing routine is briefly described along the data generation in \ref{app:data_generation_KS}. The smaller the variance in the validation set scores, the more robust a model is against weight initialization and training data distribution.

\section{Results}\label{sec:results}
We study the SSP as novel machine learning metric in regards of three central aspects: i) the property to shape gradients (convergence), ii) the regularization effect of the loss function (smoothness and final loss), and iii) compatibility with conventional optimizers (instability) such as Adam \citep{Kingma.2014} and SGD \citep{Robbins1951, Kiefer1952}. The SSP is chosen as the baseline metric for assessing the similarity between the ground truth $\mathbf{y}$ and prediction $\hat{\mathbf{y}}$. All validation and prediction errors presented hereafter hence are computed in terms of SSP, even if the individual models were trained towards a different loss function. We are well aware that this choice can seem biased or doubtful to the reader. For this reason, we will consult the well-established MSE metric for reference alongside the SSP as a validation set metric. Under outermost neutrality, we are convinced that the SSP is superior to Euclidean metrics for comparing two signals that are smooth and oscillatory following the reasoning presented before. 

We introduce the convention of stating the model training loss function as a subscript of the evaluation metric, e.g. $\mathrm{SSP}_{\mathrm{MSE}}$ indicates the validation set error measured in terms of SSP for a model trained with MSE loss function. As the loss function shapes the gradient used for training, a model trained on the MSE loss is, naively, expected to minimize the MSE validation loss best, while a model trained on the SSP loss is expected to arrive at the smallest SSP validation loss.

\subsection{Chaotic dynamics in the 1D KS equation}\label{sec:results_ks1d}
\begin{figure}[h!]
    \centering
    \includegraphics[scale=1]{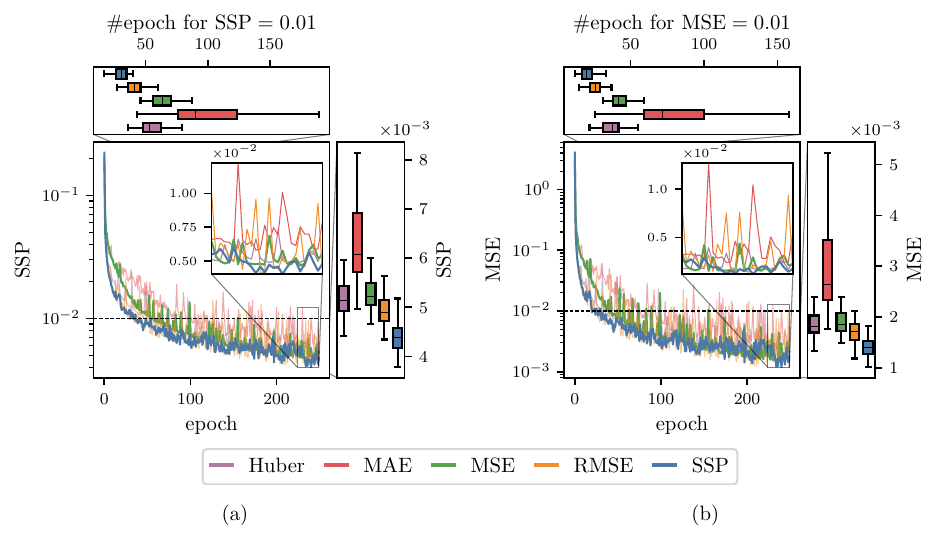}
    \caption{Model training on data of the 1D KS-eq. ($\nu=0.25$, $\Delta t = 0.1$\,s) evaluated per (a) SSP metric and (b) MSE metric. 50 training runs on random data splits are performed for each loss. The loss curves show the validation loss of one exemplary but representative training run. The vertical box-and-whisker plots show the distribution of the minimum validation loss over the last 25 epochs across all 50 training runs. The horizontal box-and-whisker plots above each loss curve show the distribution of the epoch count at which the loss falls below the threshold $\mathrm{SSP}=0.01$ ($\mathrm{MSE}=0.01$, indicated by the horizontal dashed line)}
    \label{fig:ks_1d_ssp_vs_rmse}
\end{figure}

The training behavior of machine learning models is usually evaluated using the loss curve, which provides information about the final loss value a model arrives at, the speed of convergence and the overall data fitting. We present and discuss exemplary validation loss curves for all loss functions in the following. Plotting the average loss curves over all 50 training runs would smooth out certain characteristics, hence Figure~\ref{fig:ks_1d_ssp_vs_rmse} shows a representative validation loss curve per loss function and evaluation metric. Particularly, loss curves are shown for training runs on the exact same data split. Model performance on the validation set is evaluated in terms of the SSP (Figure~\ref{fig:ks_1d_ssp_vs_rmse}~(a)) and MSE metric (Figure~\ref{fig:ks_1d_ssp_vs_rmse}~(b)) after each epoch. Box plots in Figure~\ref{fig:ks_1d_ssp_vs_rmse} depict summary statistics derived from all $50$ training runs per loss function. For all loss functions and all model training runs there was no overfitting observable.    

To assess the effect of the different loss functions on the training behavior, we first discuss the convergence speed. The horizontal box plots in Figure~\ref{fig:ks_1d_ssp_vs_rmse} depict the number of training epochs that was required to reach a certain loss threshold, e.g. $\text{SSP}=0.01$ and $\text{MSE}=0.01$, respectively. On average, models trained with SSP loss reached the threshold within $30$ epochs, earliest at $17$ and latest after $60$ epochs. Models trained on RMSE loss show comparable convergence speed, although the mean number of epochs is $41$. Huber loss and MSE loss show slower convergence with median of $53$ and $64$ epochs, respectively. The MAE loss shows a clearly different result with a very large variation and $90$ epochs required on average to achieve $\text{SSP}=0.01$. The variance in the model convergence behavior is minimal for the SSP and maximal for the MAE: The MAE shows the largest interquartile range (IQR) of 48 epochs. MSE and Huber both have an IQR of 15 epochs. RMSE (10 epochs) and SSP (9 epochs) have the lowest IQRs. As non-overlapping IQRs allow for statements that are statistically significant, this experiment indicates the superiority of the SSP loss function in terms of convergence speed. The qualitatively same picture is resembled when evaluating the validation set predictions in terms of MSE instead of SSP, see Figure~\ref{fig:ks_1d_ssp_vs_rmse}~(b). Notably, models trained on SSP loss will minimize the MSE-rated prediction error faster than models trained on MSE loss. This observation will be confirmed in the following experiments on different data sets. Therefore, the SSP loss function is shown to extend the features taken into account by Euclidean loss functions, specifically the phase and frequency information, which was exemplified in Section~\ref{sec:ssp_vs_euclidean}.   

Next, we focus on the final validation loss, i.e. the performance of several models trained for 250 epochs on the same data set but given a different loss function.  The vertical box-and-whisker plots in Figure~\ref{fig:ks_1d_ssp_vs_rmse} shows the distribution of the minimum loss value within epochs 225 and 250 across all training runs per loss function. The models trained with SSP loss arrive at a median validation loss of $\mathrm {SSP}_{\mathrm{SSP}}=0.0044$, and the model trained on MSE loss arrives at $\mathrm{SSP}_{\mathrm{MSE}}=0.0052$. The Huber loss performs roughly the same as the MSE, the RMSE arrives at a slightly lower final loss value. Lastly, the MAE shows the weakest performance at a median of $\mathrm{SSP}_{\mathrm{MAE}}=0.0061$. Considering the variance among all training runs, the MAE has the largest IQR of $0.0012$. The remaining loss functions exhibit similar IQRs at $0.0004$ (SSP and RMSE) and $0.0005$ (MSE and Huber). As the IQR of the SSP does not overlap with any of the other loss function's IQRs, one can state that the superiority of the SSP in terms of lowest final loss is statistically significant. Lastly, one can observe that the curves fluctuate for all loss functions. However, in direct comparison, the SSP loss function exhibits the smoothest loss characteristic. The loss curves allow to glance at the optimizer paths towards a minimum: The less fluctuating loss curve for the SSP thus indicates that the optimizer has less trouble finding it's path downwards on the loss surface compared to the other loss functions. 

To conclude, for the presented case, using the SSP loss function results in smoother loss curves which point towards better regularization properties than the competing loss functions. The SSP loss function not only speeds up the training but also results in better validation scores after a pre-defined number of training epochs. Training a model with SSP loss function results in lower MSE validation errors as when training a model with MSE loss. This behavior indicates the regularization property, i.e. shows that the SSP is penalizing small prediction errors more strongly than other loss functions.

\subsection{Gradient shaping through SSP loss function}
\begin{figure}[h!]
     \centering
     \includegraphics[scale=1]{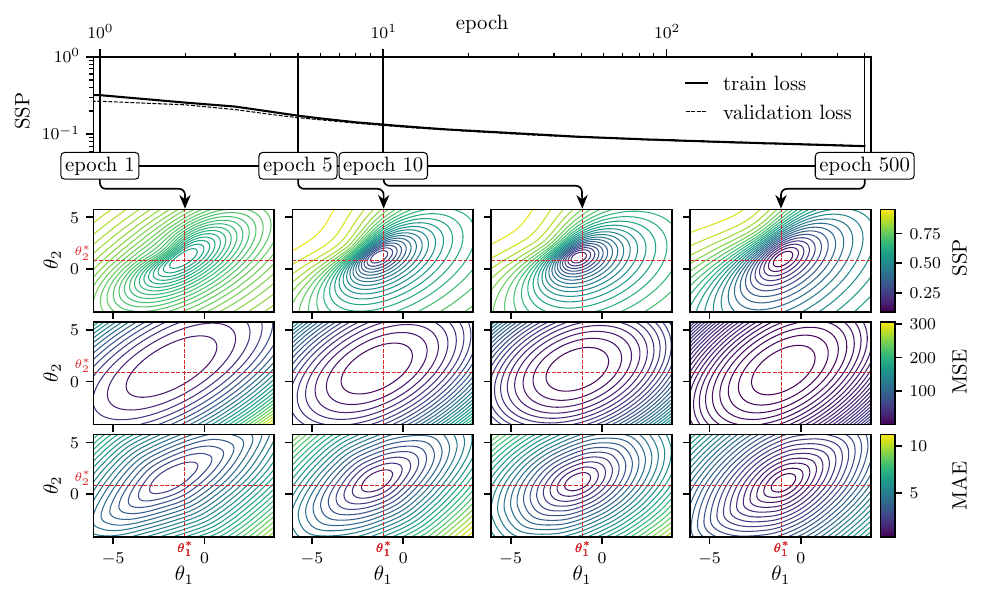}
     \caption{Visualization of loss hyperplanes for a single data sample by variation of two weights $\theta_1$,  $\theta_2$ in the last layer of the model about the final weights $\theta_1^*$, $\theta_2^*$ reached after 500 epochs. Loss hyperplanes are displayed for four stages of the training, and model predictions are evaluated in terms of SSP (top row), MSE (middle row) and MAE (bottom row) for the 1D KS-eq. ($\nu=0.25$, $\Delta t=1.0\,\mathrm{s}$). The overall training process is depicted in the top panel}
     \label{fig:loss_hyperplane}
\end{figure}

The training behavior discussed in the previous paragraph indicates several advantages of the SSP loss over other loss functions for the given prediction task. In order to understand some of those observations, and to come back to the discussion in Section~\ref{sec:ssp_vs_euclidean}, Figure~\ref{fig:loss_hyperplane} displays the loss hyperplane of a single model (1D KS-eq., $\nu=0.25$, $\Delta t=1.0\,\mathrm{s}$, SSP loss) at different stages of the training process, namely after epochs 1, 5, 10, and 500. For this analysis, two weights in the last layer, $\theta_1^*$ and $\theta_2^*$, are varied while all other weights are kept frozen. The prediction of that model for a single data sample is performed for a wide variation of $\theta_1, \theta_2$ about the final weights $\theta_1^*=-1.07$, $\theta_2^*=0.86$ reached after 500 epochs of training. For every choice of the weight values, the model prediction is measured in terms of SSP, MSE, and MAE metric. As a result, one can study the loss hyperplane and its gradients that the optimizer would see for that single data sample if only two weights of the model were trainable. Even though this situation does not reflect the actual training process, the error surfaces depict salient differences that can link to the observations made before on the training behavior of the SSP loss. The error isolines of the MAE are separated by a constant value, indicating a linear gradient towards the minimum. The MSE shows a different behavior, which resembles the shape of a flat bowl as gradients get smaller towards the optimum. The SSP error surface is not as circular, and exhibits large values particularly in the upper left parameter region. Gradients are very large in the anti-diagonal direction of the parameter space. The most important difference in the three metrics lies in the area surrounding the minimum at $(\theta_1^{*}, \theta_2^{*})$: while the MSE would allow for a significant change in the weights without crossing the lowest isoline, the SSP would increase very quickly under small variations of the weights. For example, reducing $\theta_1^*$ by $10\%$ will increase the MSE by $0.3\%$, the MAE by $2.1\%$, and the SSP by $56.5\%$, in relation to the full range of the respective hyperplane, for the model after epoch 10. While the Euclidean metrics may accept a small variation in that weight, the SSP would respond with a severe increase of the loss value. Concluding, the local minima in the illustrated SSP error surface are very sharp and narrow, therefore the training process may converge faster and more precisely to the minimum, as the SSP is not as lenient to weight changes as the MAE and MSE losses are. While we were able to reproduce the preceding result by changing randomly picked sets of weights from the last layer, it is not feasible to illustrate the loss surface for all parameters. Here, we want to point towards the results for the actual training (i.e. all parameters are involved) presented in Section~\ref{sec:results_ks1d} to statistically support our hypothesis.

\subsection{Effect of prediction task complexity}
This section displays several experiments on the one- and two-dimensional KS-eq. for different values of $\nu$, i.e. different levels of chaos in the physical problem. Further, the time step $\Delta t$ is varied, which denotes the time difference between consecutive data samples used as input to the flow map model. Hence, for smaller $\nu$ and for larger $\Delta t$ the learning task becomes more demanding, as more chaotic dynamics are to be predicted over longer time instants into the future. The role of small perturbations and inaccuracies is increased according to the exponential divergence driven by positive Lyapunov spectra. Table~\ref{tab:overview} displays validation scores for five KS-eq. cases of increasing complexity. For each case, models are trained on five different loss functions, and the validation set scores are evaluated in terms of the SSP and MSE metric. Cross-validation is performed against 50 realizations (10 for the 2D cases) of each experiment.
\begin{table}[h!]
    \caption{Validation scores for all case studies, sorted by the estimated complexity of the prediction task from top (low complexity) to bottom (high complexity). For a given model and the same data set, training is performed for 250 epochs (100 for KS-2D cases) using different loss functions. In each case, the model prediction quality on the validation set is measured by metrics SSP and MSE. 50 training runs (10 for KS-2D) are performed for each set-up on different data splits, and mean $\pm$ standard deviation of validation scores are reported. Bold font and green backgrounds highlight the best score per metric and case}
    \label{tab:overview}
    \centering
    \begin{tabular}{l l c c}
    \toprule
        case & loss & metric SSP & metric MSE\\
        \midrule
        \addlinespace
         KS-1D ($\nu = 1.0,\;\Delta t = 0.1\,\mathrm{s}$)
         & SSP   & $0.0034\pm 0.0008$ & $0.0006\pm 0.0004$\\
         & MSE   & $0.0023\pm 0.0003$ & $0.0002\pm 0.0001$\\
         & MAE   & $0.0045\pm 0.0012$ & $0.0016\pm 0.0030$\\
         & RMSE  & $0.0030\pm 0.0006$ & $0.0005\pm 0.0004$\\
         & \cellcolor{green!25}\textbf{Huber} & \cellcolor{green!25}$\mathbf{0.0022\pm 0.0002}$ & \cellcolor{green!25}$\mathbf{0.0002\pm 0.0000}$\\
         \midrule
        \addlinespace
         KS-1D ($\nu = 0.25,\;\Delta t = 0.1\,\mathrm{s}$)
         & \cellcolor{green!25}\textbf{SSP}   & \cellcolor{green!25}$\mathbf{0.0044\pm 0.0003}$ & \cellcolor{green!25}$\mathbf{0.0014\pm 0.0002}$\\
         & MSE   & $0.0053\pm 0.0004$ & $0.0020\pm 0.0004$\\
         & MAE   & $0.0063\pm 0.0010$ & $0.0034\pm 0.0021$\\
         & RMSE  & $0.0049\pm 0.0004$ & $0.0018\pm 0.0006$\\
         & Huber & $0.0052\pm 0.0004$ & $0.0018\pm 0.0003$\\
         \midrule
         \addlinespace
         KS-1D ($\nu = 0.25,\;\Delta t = 1.0\,\mathrm{s}$)
         & \cellcolor{green!25}\textbf{SSP}   & \cellcolor{green!25}$\mathbf{0.0755\pm 0.0020}$ & \cellcolor{green!25}$\mathbf{0.4554\pm 0.0210}$\\
         & MSE   & $0.0800\pm 0.0023$ & $0.4882\pm 0.0262$\\
         & MAE   & $0.0812\pm 0.0021$ & $0.5039\pm 0.0253$\\
         & RMSE  & $0.0781\pm 0.0023$ & $0.4728\pm 0.0247$\\
         & Huber & $0.0806\pm 0.0023$ & $0.4958\pm 0.0243$\\
         \midrule
         \addlinespace
         KS-2D ($\nu=0.25,\;\Delta t=0.1\,\mathrm{s}$) 
         & \cellcolor{green!25}\textbf{SSP}   & \cellcolor{green!25}$\mathbf{0.0074\pm0.0003}$ & \cellcolor{green!25}$\mathbf{0.0043\pm0.0004}$\\
         & MSE   & $0.0090\pm0.0004$ & $0.0061\pm0.0006$\\
         & MAE   & $0.0091\pm0.0005$ & $0.0065\pm0.0008$\\
         & RMSE  & $0.0090\pm0.0011$ & $0.0064\pm0.0018$\\
         & Huber & $0.0089\pm0.0008$ & $0.0064\pm0.0019$\\
         \midrule
         \addlinespace
         KS-2D ($\nu=0.25,\;\Delta t=1.0\,\mathrm{s}$) 
         & \cellcolor{green!25}\textbf{SSP} & \cellcolor{green!25}$\mathbf{0.2336\pm 0.0044}$ & \cellcolor{green!25}$\mathbf{4.2762\pm 0.0833}$\\
         & MSE & $0.2483\pm 0.0033$ & $4.2998\pm 0.0619$\\
         & MAE & $0.2473\pm 0.0025$ & $4.3478\pm 0.0347$\\
         & RMSE & $0.2484\pm 0.0035$ & $4.3055\pm 0.0848$\\
         & Huber & $0.2470\pm 0.0028$ & $4.3220\pm 0.0743$\\
         \midrule
         \addlinespace
         dispersive waves ($\Delta t=1.3\,\mathrm{s}$)&
          \cellcolor{green!25}\textbf{SSP}   & \cellcolor{green!25}$\mathbf{0.0162\pm0.0015}$ & \cellcolor{green!25}$\mathbf{0.0005\pm0.0001}$\\
        & MSE   & $0.0210\pm0.0018$ & $0.0008\pm0.0001$\\
        & MAE   & $0.0242\pm0.0016$ & $0.0010\pm0.0001$\\
        & RMSE  & $0.0190\pm0.0016$ & $0.0007\pm0.0001$\\
        & Huber & $0.0209\pm0.0017$ & $0.0008\pm0.0001$\\
         \bottomrule
    \end{tabular}
\end{table}

The one-dimensional KS-eq. case using $\nu=1.0$, $\Delta t=0.1$\,s represents the seemingly simplest prediction task. Here, models trained on Huber loss outperform models trained on any other loss function after the same amount of training epochs. Huber loss reduces both validation metrics the most, and also across all 50 different models trained on different splits of the data set. The SSP loss achieves weaker results, while the MSE loss achieves only slightly weaker results than the Huber loss. Moreover, the range of validation scores in the cross validation study is minimal for the Huber loss models. With increasing prediction task complexity ($\nu=0.25$, $\Delta t = \{0.1, 1.0\}\,\mathrm{s}$), we observe an expected degradation in overall prediction quality for all models. However, both experiments show a similar result: models trained on the SSP loss achieve the best validation scores, even in terms of MSE. That is, if a model is trained using SSP loss, it outperforms a model trained on MSE, even if the validation score is measured in terms of MSE. Moreover, the standard deviation of validation scores across all 50 models trained on SSP loss is the smallest for both experiments and both validation metrics across all loss functions. For the two-dimensional KS-eq. cases, a diverse observation is made: using a time increment of $\Delta t = 0.1$\,s, no performance degradation is visible, and even an increase in prediction accuracy can be read from the SSP metric when compared to the one-dimensional case studies. However, as the time increment is increased, the prediction task is rendered much more demanding as with larger time steps, the effect of nonlinearities will become more prominent such that the next system state cannot be estimated by linear extrapolation from the current system state. As a result, considerable performance degradation is observed for $\Delta t=1.0\,\mathrm{s}$ throughout all loss functions. Still, the SSP loss function manages to minimize the loss in terms of both SSP and MSE metric best. The remaining loss functions perform on par with each other, showing only minor deviations of $\Delta_{\mathrm{SSP}}\leq1.3\times10^{-3}$ or $\Delta_{\mathrm{MSE}}\leq4.8\times10^{-2}$ between results.

The experiments conducted on the Kuramoto-Sivashinsky equation indicate that the SSP is especially well-suited for more complex prediction tasks on sequential data with an oscillatory character. While the Huber loss performs best on the least complex problem, the models trained with SSP loss achieve better validation scores for the remaining problems. The 2D implementation of the SSP manages to pull ahead from the competition. Most importantly, the SSP loss function manages to reduce the loss in terms of MSE better than the MSE itself for all cases except the least complex one. While the presented results are achieved using the Adam optimizer \citep{Kingma.2014}, we ran a subset of the experiments listed in Table~\ref{tab:overview} using the SGD optimizer \citep{Robbins1951, Kiefer1952}. The loss curves for the SSP showed a very smooth convergence compared to the remaining loss functions, however training turned out to be unstable in some cases across all loss functions. Overall, the results on SGD, while seeming less stable, confirm the observations presented here.

\subsection{Dispersive wave propagation}
\begin{figure}[h!]
    \centering
    \includegraphics[scale=1]{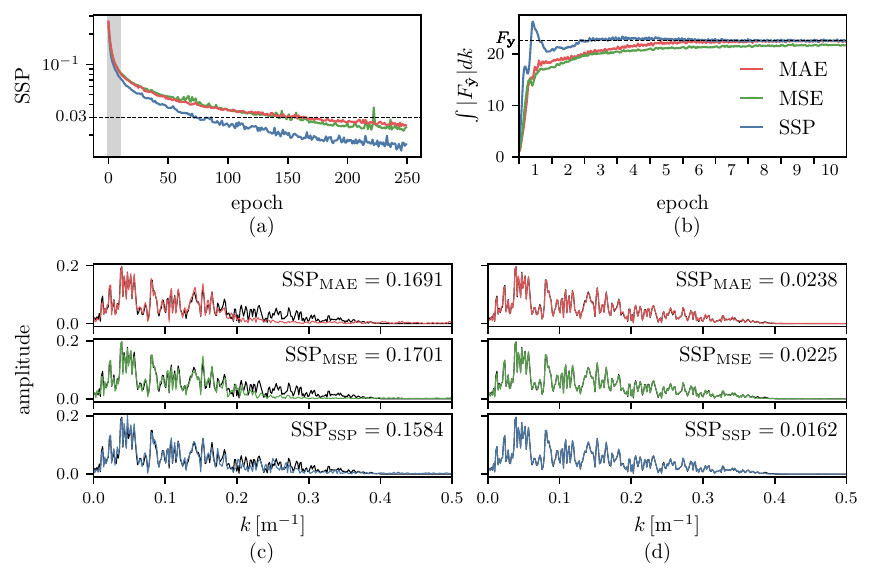}
    \caption{Model training on the 1D dispersive wave data evaluated in terms of SSP metric. The loss curves (a) show the validation loss for models trained with SSP, MSE and MAE loss functions. The gray area marks the first ten epochs, where we obtain predictions for one validation sample after each weight update. (b) shows the integral of the Fourier spectra of the respective prediction after each weight update. The integral of the ground truth is indicated by the dashed line. Panels (c) and (d) show the Fourier spectra of the predictions using the model trained with MAE, MSE and SSP loss functions (from top to bottom) after 2 (c) and 250 (d) training epochs, respectively. The Fourier spectrum of each prediction is evaluated against the ground truth spectrum shown in black}
    \label{fig:disp_waves_overview}
\end{figure}

The third evaluation case deals with one-dimensional dispersive surface waves governed by the linear wave equation Eq.~\eqref{eq:disp_waves}. Compared to the one- and two-dimensional nonlinear KS-eq., the surface waves are characterized by a very broad wave spectrum, rendering this prediction case conceptually different to the KS-eq. experiments. The dispersion relation lets long waves travel faster than short waves, hence creating complex spatio-temporal patterns. Details on the data generation are presented in \ref{app:data_generation_disp_waves}. In analogy to the 1D KS-eq., we cross validate our results against 50 training runs on independent and random data splits, and each model is trained for 250 epochs. Figure~\ref{fig:disp_waves_overview}~(a) shows three exemplary model trainings using MAE, MSE and SSP as loss function. Here, we can observe that the loss curves for MAE and MSE overlap for the most part while the model with SSP loss undershoots both MAE and MSE for all epochs. The validation scores (see Table~\ref{tab:overview}) indicate that models with SSP loss outperform models using different loss functions on the dispersive wave data. On average, a mean validation set score of $\mathrm{SSP}_{\mathrm{SSP}}=0.0162$ is achieved. The models trained on RMSE performs second best ($\mathrm{SSP}_{\mathrm{RMSE}}=0.0190$), MSE and Huber loss perform equally well at a mean $\mathrm{SSP}\approx0.0210$ and lastly the MAE performs the weakest ($\mathrm{SSP}_{\mathrm{MAE}}=0.0242$).

To quantify the convergence speed of each loss function, we define a threshold at $\mathrm{SSP}=0.03$, indicated by the dashed line in Figure~\ref{fig:disp_waves_overview}~(a). On average over all 50 training runs, models with MAE loss reach the threshold within 142 epochs, and models with MSE take 130 epochs. Models trained on RMSE and Huber loss take 108 and 120 epochs, respectively. The models trained with SSP loss converge considerably faster and reach the threshold after 78 epochs. Training on Huber and SSP loss shows the smallest variability among the loss functions with an IQR of 16 and 17 epochs, respectively. MSE and RMSE model trainings both have an IQR of 24 epochs while the MAE loss exhibits the largest variability with an IQR of 34 epochs. Variability is directly linked with the sensitivity of the training process against weight initialization and data distribution in the training sets. The low training variability in combination with the highest convergence speed makes the SSP loss function superior to the competition in this experiment. Again, the difference in convergence speed can be stated statistically significant based on the non-overlapping IQRs.

To shed more light onto the training process, we use the model to predict one validation sample after each weight update within the first 10 epochs of training (using a batch size of 2048, we obtain 81 weight updates per epoch). The results discussed hereafter are representative for a wide range of sample signals that were analyzed. In addition, both RMSE and Huber, being Euclidean distance-based metrics, exhibit the same characteristics as portrayed by MSE and MAE and are therefore not presented explicitly. Figures~\ref{fig:disp_waves_overview}~(c) and (d) show the spectra of the predictions after 2 and 250 epochs of training, respectively. After 2 epochs, the models trained on MSE and MAE achieve validation scores around $\mathrm{SSP}\approx0.17$ while the model trained on SSP achieves a slightly lower validation score of $\mathrm{SSP}_{\mathrm{SSP}}=0.1584$. The spectra indicate that this marginal lead is due to the fact that the model trained with SSP already attempts to fit higher frequency components while the spectra associated with MAE and MSE remain mostly flat for $k>0.2\,\mathrm{m}^{-1}$. After 250 epochs all models fit the ground truth spectrum well over all frequencies with a slight advantage of the SSP over MSE and MAE. Figure~\ref{fig:disp_waves_overview}~(b) shows the integral of the Fourier transform of each prediction compared against the ground truth integral value. As the integral represents a measure of overall energy contained in the predicted wave signal, this value is mostly related with amplitude errors. While the models trained with MAE and MSE loss functions slowly approach the ground truth value, the model trained with SSP exhibits an overshooting behavior in the very first training epoch. Hereafter, the SSP model quickly approaches the ground truth. These observations link well with the analytical discussion of the SSP metric and the conceptual analysis presented in Section~\ref{sec:ssp_vs_euclidean}. First, as conventional Euclidean error metrics are amplitude-biased, the overall energy in the predictions is very small during the first training epochs. To reduce the loss value, the optimizer reduces the overall amplitudes first, before energy is slowly increased to match the ground truth. On the contrary, the SSP is not amplitude-biased, such that also a frequency or phase error can increase the loss value. The above mentioned characteristic of the SSP to shape higher frequency components during early training results from this property. Second, the convergence speed towards the ground truth value is linked to the gradients in vicinity of a loss minimum: the MSE gradient becomes smaller as the minimum is approached, such that convergence becomes slower over the number of training epochs. Even for excessive training, the MSE exhibits a non-vanishing difference to the ground truth, which is fundamentally rooted in the bowl-type shape of the MSE. The constant gradient of the MAE results in a steady approach of the minimum. Conceptually different, the gradients of the SSP are very sharp in vicinity of a minimum, hence increasing convergence speed. Predictions far from the optimum enter the plateau in the loss surface that exists in sub-optimal regions of the parameter space. Overall, the SSP is the most unforgiving in cases of small errors in amplitude, phase, or frequency, hence enhancing the convergence behavior of the neural network training. This supremacy becomes especially evident for the broad-banded wave spectrum, while it is already observable for the chaotic KS-eq. case studies. Thus, we conclude that the SSP is a particularly promising alternative to conventional loss functions when training machine learning models on sequential data with complex spectra.

\section{Conclusions}\label{sec:conclusion}
This work proposes the surface similarity parameter (SSP) as a novel machine learning loss function that is designed for oscillatory sequences arising across many fields of research related to physics and natural processes. Our experiments cover the chaotic dynamics in the one- and two-dimensional Kuramoto-Sivashinsky equation and dispersive surface gravity waves in fluids.

Our research investigates the effect of the loss function on the training behavior of a machine learning flow map, i.e. a fully convolutional neural network that advances historic state observations one time increment into the future. The experiments indicate that the proposed SSP loss function is superior to classical Euclidean loss functions, such as MSE and MAE, in terms of a) convergence rate, b) robustness against weight initialization effects, and c) final model prediction accuracy. Compared to conventional Euclidean loss functions, the SSP loss function is a bounded error metric that exhibits large gradients in vicinity of local minima. This regularization property allows to faster train more precise models. While Euclidean distance-based metrics turn out to be amplitude-biased for oscillatory data, the SSP metric is taking amplitude, phase and frequency information explicitly into account, and is therefore much stricter even for small deviations between prediction and ground truth sequences. The superiority of the SSP is especially evident the more complex the spectral composition of the sequential data is, i.e. the wider the spectrum of harmonic components and the more irregular the amplitude distribution becomes. The novel SSP loss function is therefore highly promising for regression on oscillatory sequential data.

\clearpage
\appendix

\section{Conventional loss functions}\label{app:loss_functions}
\setcounter{table}{0}
Table~\ref{tab:loss_overview} lists an overview on all conventional Euclidean loss functions used in the course of this work, accompanied by a short description of the interpretation. 

\begin{table}[h]
    \centering
        \caption{Overview on common regression loss functions for sequence prediction tasks}
            \label{tab:loss_overview}
    \begin{tabular}{p{2.5cm} p{6cm} p{5.5cm}}
    \toprule
    loss & definition & comment \\
    \midrule
        mean absolute error & $\mathrm{MAE} = \frac{1}{N} \sum_{i}^{N} \left|y_i - \hat{y}_i \right|$  &  arithmetic mean of absolute errors, also denoted $L$1 loss. Estimates the median of the error distribution function, therefore robust against outliers. But: constant and large (!) gradients also for small errors, requires adaptive learning rates for better convergence. Gradients are not continuous (not differentiable at 0.)\\[2.0ex]
        
        mean squared error & $\mathrm{MSE} = \frac{1}{N} \sum_{i}^{N} \left(y_i - \hat{y}_i \right)^2$ & arithmetic mean of squared errors, also denoted $L$2 loss. Estimates the mean of the error distribution function, put emphasis on large values. Large gradients at large errors, small gradients at small errors. Helps training neural networks with constant learning rate \\[2.0ex]
        
        root mean squared error & $\mathrm{RMSE} = \sqrt{\frac{1}{N}\sum_{i}^{N} \left(y_i - \hat{y}_i \right)^2}$ & arithmetic mean of squared errors, also denoted $L$2 loss. Same scale as MAE. Estimates the mean of the error distribution function, put emphasis on large values.  \\[2.0ex]
        
        Huber loss & $\mathcal{L}_{\delta} = \begin{cases} \frac{1}{2}(y - \hat{y})^2 & \mbox{for } |y - \hat{y}|\leq\delta, \\\delta(|y - \hat{y}|-\frac{1}{2}\delta) & \mbox{else}\end{cases}$ & Combination of MSE (small errors) and MAE (large errors) with small and large gradients, respectively, and differentiable at 0.\\
         \bottomrule
    \end{tabular}
\end{table}

\section{On the optimizer step size}\label{sec:toy_example}
\setcounter{figure}{0}
\setcounter{table}{0}
The optimizer used for ML model training has the crucial task of finding a local minimum in the loss surface. Purely gradient-based optimization schemes like SGD \citep{Robbins1951, Kiefer1952} rely on large gradients, as the optimizer's step size is directly linked to the gradient, $\boldsymbol{\theta}_{t+1}=\boldsymbol{\theta}_t - \alpha\nabla_{\boldsymbol{\theta}}J(\boldsymbol{\theta}_t)$ with model parameters $\boldsymbol{\theta}$, learning rate $\alpha$ and iteration count $t$. Adaptive momentum estimation (Adam) \cite{Kingma.2014} injects $1^{\text{st}}$ and $2^{\text{nd}}$ order momentum, $\hat{\boldsymbol{m}}_t\left(\beta_1, \nabla_{\boldsymbol{\theta}}J(\boldsymbol{\theta})\right)$ and $\hat{\boldsymbol{v}}_t\left(\beta_2, \nabla_{\boldsymbol{\theta}}^2J(\boldsymbol{\theta})\right)$ with $1^{\text{st}}$ and $2^{\text{nd}}$ order exponential decay $\beta_1$ and $\beta_2$, into the weight update scheme, $\boldsymbol{\theta}_{t+1} = \boldsymbol{\theta}_t - \alpha\tfrac{\hat{\boldsymbol{m}}_t}{\sqrt{\hat{\boldsymbol{v}}_t} - \varepsilon}$. Thus, the step size in parameter space does not only rely on the current but also on past gradients. \cite{Kingma.2014} state that the step size of Adam has an upper bound ``when a gradient has been zero at all time steps except the current one." To evaluate the effect of the loss surface on the step size of the Adam optimizer ($\alpha=0.001,\;\beta_1=0.9,\;\beta_2=0.999,\;\varepsilon=1\mathrm{e}{-8}$), we introduce a one dimensional toy-example using two potentials
\begin{linenomath*}
\begin{align}
    J_{\text{(a)}}(\theta) &= 1 - \cosh^{-2}(\theta)\in[0,1], & J_{\text{(b)}}(\theta) &= \left(\frac{\theta}{\theta_0}\right)^2\in\mathbb{R}^+,\quad\theta_0 = -3
\end{align}
\end{linenomath*}
of which $J_{\mathrm{(a)}}$ is characterized by a sharp minimum and thus qualitatively resembles the loss surface of SSP, while $J_{\mathrm{(b)}}$ qualitatively resembles the bowl-shaped loss surface of the MSE. $J_{\mathrm{(b)}}$ is normalized with the initial weight $\theta_0$, such that $J_{\mathrm{(b)}}(\theta_0)=1$. Both potentials have their minimum at $\theta^\star=0$.
\begin{figure}
    \centering
    \includegraphics{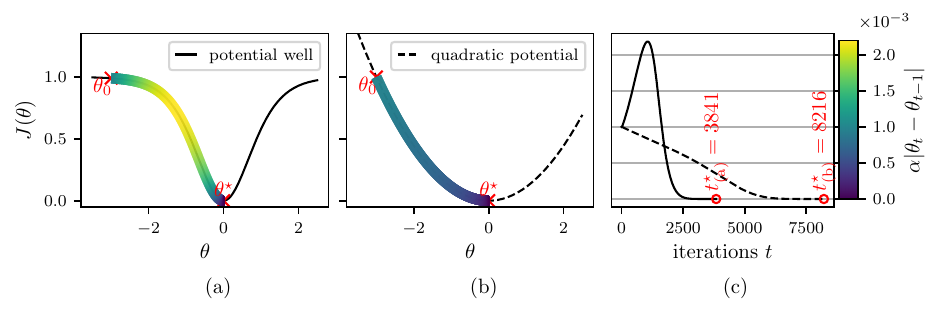}
    \caption{Convergence speed of Adam optimization scheme on (a) potential well and (b) quadratic potential. Both potential functions have their minimum at $\theta^\star=0$, the initial weight is $\theta_0=-3,\;J_{\text{(a)}}(\theta_0)=0.99,\;J_{\text{(b)}}(\theta_0)=1$. (c) shows the optimizer step size $\Delta\theta=\alpha|\theta_t - \theta_{t-1}|$ over the iteration count $t$. The step sizes are also superimposed onto the potential functions as multicolored lines. The Adam optimization takes $t_{\text{(a)}}^\star=3841$ iterations on the potential well (a) and $t_{\text{(b)}}^\star=8216$ iterations on the quadratic potential (b)}
    \label{fig:toy_example_adam}
\end{figure}
We define the initial weight $\theta_0=-3$, such that
\begin{linenomath*}
\begin{equation}
	\left(J_{\mathrm{(a)}}(\theta_0)\approx J_{\mathrm{(b)}}(\theta_0)\right)\land\left(\left|\nabla_{\theta}J_{\mathrm{(a)}}(\theta_0)\right| \ll\left| \nabla_{\theta}J_{\mathrm{(b)}}(\theta_0)\right|\right)\text{.}
\end{equation}
\end{linenomath*}
Each optimizer is applied to the potentials $J_{\text{(a)}}$ and $J_{\text{(b)}}$ and step sizes $\alpha|\theta_t - \theta_{t-1}|$ over the iteration count $t$ are logged. $t^\star$ resembles the iteration count it takes for the algorithm to converge to $|\theta - \theta^\star|<1\mathrm{e}{-8}$. While the step size for the quadratic potential is steadily descending over $t$, the step size for the potential well increases to more than double its initial value and then falls off quickly. Figure~\ref{fig:toy_example_adam}~(a) indicates that the maximum step size is reached shortly after the well begins, i.e. the initial plateau in $J_{\text{(a)}}$ causes the optimizer to ramp up the step size, while the convex shape of $J_{\text{(b)}}$ leads to a steadily decreasing step size. Convergence is faster for the potential well than for the quadratic potential, which were chosen to resemble the SSP and MSE loss surfaces, respectively.

We study SGD ($\alpha=0.01$, no momentum) under identical initial conditions. Since SGD is purely gradient-based, the step sizes can be directly derived from the potential shape. Therefore, the initial step size in $J_{\text{(b)}}$ is significantly larger than in $J_{\text{(a)}}$. However, as $\theta$ converges towards the optimum $\theta^\star$, $\nabla_{\theta}J_{\text{(b)}}\to 0$, leading to vanishing step sizes for the quadratic potential. For $J_{\text{(a)}}$, being first concave and then convex, SGD starts with a very small step size which is then greatly increased and reaches its maximum in the inflection point of $J_{\text{(a)}}$ (the maximum of $\nabla_{\theta}J_{\text{(a)}}$). SGD converges after $3539$ iterations for the potential well, and after $8774$ iterations for the quadratic potential.

\section{Data generation}\label{app:data_generation}

\subsection{Kuramoto-Sivashinsky equation}\label{app:data_generation_KS}
The Kuramoto-Sivashinsky equation has been studied in several recent works \citep{Smaoui2004, Pathak2017, Pathak.2018, Raissi.2018, vlachas2018, BarSinai2019} related to machine learning as an illustrative example for complex spatio-temporal dynamics. For our case studies, we use the KS-eq.
\begin{linenomath*}
\begin{equation}\label{eq:KS_equation}
    u_t + \Delta u + \nu\Delta^2 u + \frac{1}{2}|\nabla u|^2
\end{equation}
\end{linenomath*}
augmented with a viscosity term $\nu\in\{0.25, 1\}$ that controls the amount of chaos in the system \citep{Smaoui2004, vlachas2018}. We initialize both our 1D and 2D KS systems with the superposition of 10 waves
\begin{linenomath*}
\begin{equation}\label{eq:KS_initial_conditions}
    u(x) = \sum\limits_{i=1}^{10} A_i\sin(2\pi\frac{i}{L}x + \varphi_i), \quad
    u(x,y) = \sum\limits_{i=1}^{10} A_i\sin(2\pi\frac{i}{L}\left(x\cos(\phi_i) + y\sin(\phi_i) \right) + \varphi_i)
\end{equation}
\end{linenomath*}
with random amplitude $A_i\in[-1, 1]$ and random phase $\varphi_i\in[0, 2\pi]$, $\phi_i\in[-\pi/2, \pi/2]$. For the 1D KS equation, we use a domain $x$ with length $L_x=128\,\mathrm{m}$ discretized into 1024 grid points. The 2D KS equation uses a square domain with the same set-up as the 1D case in $x$ and $y$ dimension. The definition of our initial conditions Eq.~\eqref{eq:KS_initial_conditions} ensure that the longest wave fits the domain exactly. All derivatives in Eq.~\eqref{eq:KS_equation} are calculated using the 4th order finite differences method. To ensure numeric stability, we apply a low pass filter so that only the first 80 (2D: 60) frequency components are taken into account after each time step. For time integration of the PDE we use the modified Euler method \citep{Sueli2003, Burden2010}. The system is simulated using $\Delta t_{\mathrm{sim}}=0.01\,\mathrm{s}$ of which every 10th system state is saved to disk, resulting in a data time step size of $\Delta t=0.1\,\mathrm{s}$. The KS-eq.~\eqref{eq:KS_equation} is simulated 100 times (2D: 10 times) for $200\,\mathrm{s}$ for $\nu=\{0.25, 1.0\}$. The simulation data is preprocessed using a sliding window of size $n_{\tau}=8$ along the time dimension with stride 1 and shift 1 for $\Delta t=0.1\,\mathrm{s}$ (shift 10 for $\Delta t=1.0\,\mathrm{s}$). For $\Delta t=0.1\,\mathrm{s}$ we obtain 1993, for $\Delta t=1.0\,\mathrm{s}$ we end up with 1921 input-output-pairs $(\mathbf{X}\in\mathbb{R}^{1024\times8},\;\mathbf{y}\in\mathbb{R}^{1024})$ for each $\nu\in\{0.25, 1.0\}$ (2D: 1993 input-output pairs $(\mathbf{X}\in\mathbb{R}^{1024\times1024\times8},\;\mathbf{y}\in\mathbb{R}^{1024\times1024})$ for $\Delta t=0.1\,\mathrm{s}$, 1921 for $\Delta t=1.0\,\mathrm{s}$).

\subsection{Dispersive surface waves}\label{app:data_generation_disp_waves}
The dispersive surface wave data is generated using linear wave theory, i.e. no interactions between waves are modeled. A superposition of $N=651$ single harmonic waves forms the initial surface wave
\begin{linenomath*}
\begin{equation}\label{eq:disp_waves}
    \eta_0 = \eta(x, t=0) = \sum\limits_{i=1}^N A_i\cos(k_i x + \varphi_i)
\end{equation}
\end{linenomath*}
with amplitude $A_i$, wave number $k_i$ and random phase $\varphi_i\in[0, 2\pi]$. The amplitudes are defined by a normalized Fourier spectrum \citep{Klein2021}
\begin{linenomath*}
\begin{equation}
    F(\omega_i) = \frac{27\left(\omega_i - \omega_{\mathrm{min}}\right)\cdot\left(\omega_i - \omega_{\mathrm{max}}\right)^2}{4\left(\omega_{\mathrm{max}} - \omega_{\mathrm{min}}\right)^3}
\end{equation}
\end{linenomath*}
with $\omega_{\mathrm{min}}=0.3\,\mathrm{rad}\,\mathrm{s}^{-1}$ and $\omega_{\mathrm{max}}=2.0\,\mathrm{rad}\,\mathrm{s}^{-1}$ limiting the relevant frequency range. The wave number $k$ is linked to the angular frequency $\omega$ via the linear dispersion relation
\begin{linenomath*}
\begin{equation}\label{eq:quadratic_dispersion}
    \omega(k) = \sqrt{g\cdot k\cdot\tanh(k\cdot d)}
\end{equation}
\end{linenomath*}
with gravitational acceleration $g$ and water depth $d=500\,\mathrm{m}$. In simple terms, the dispersion relation Eq.~\eqref{eq:quadratic_dispersion} defines the propagation velocity of a wave depending on its wave length. In consequence, long waves travel faster than short waves. To simulate the time evolution of the initial wave Eq.~\eqref{eq:disp_waves}, the assumption of linear wave theory allows to simulate the time evolution of each single harmonic wave according to their respective dispersion relation and recreate the surface wave at any given time by superposition of all single harmonic waves. Since $\eta_0$ carries all information about the space domain (wave numbers $\mathbf{k}\in\mathbb{R}^N$ and initial phases $\boldsymbol{\varphi}\in\mathbb{R}^N$), we can simulate the time evolution of $\eta_0$ in the complex space by application of the Fourier transform
\begin{linenomath*}
\begin{equation}
    F_{\eta(t)} = F_{\eta_0}\exp(-i\boldsymbol{\omega} t)
\end{equation}
\end{linenomath*}
and obtain the surface wave $\eta(t)$ by means of the inverse Fourier transform of $F_{\eta(t)}$. As this FFT-based simulation approach introduces periodic boundary conditions, the simulated wave exits the domain to the right and re-enters it on the left. To counteract this behavior, we append zeros to $\eta_0$ thus expanding the domain to the right. The initial wave now travels into the appended zeros instead of re-entering the domain. Overall, we generate 100 initial sea states with random phase $\varphi_i$ and otherwise identical configuration. Every sea state is simulated for $250\,\mathrm{s}$ with $\Delta t=0.1\,\mathrm{s}$. The data preprocessing for ML involves the application of a sliding window with stride 1 and shift 13, increasing the time step in the final data set to $\Delta t=1.3\,\mathrm{s}$. Using this approach, we obtain 2396 input-output-pairs $(\mathbf{X}\in\mathbb{R}^{1024\times8},\;\mathbf{y}\in\mathbb{R}^{1024})$ from each simulation.

\clearpage
\section*{Acknowledgements}
This work was supported by the Deutsche Forschungsgesellschaft (DFG -- German Research Foundation) [grant number 277972093]; and Hamburg University of Technology (TUHH) [$\text{I}^3$-project ``Predicting Ship Hydrodynamics to Enable Autonomous Shipping:
Nonlinear Physics and Machine Learning"].

\bibliography{main}

\end{document}